\documentclass[journal, twoside]{IEEEtran}

\usepackage{graphicx} 
\usepackage[cmex10]{amsmath} 
\usepackage{amssymb}  
\usepackage[caption=false,font=footnotesize]{subfig}
\usepackage{algorithm}
\usepackage{algpseudocode}
\usepackage{multicol}
\usepackage{minibox}
\usepackage{color}
\usepackage{cite}
\usepackage{pstool}
\usepackage{psfrag}
\usepackage{xcolor,colortbl}
\usepackage[keeplastbox]{flushend}
\usepackage{soul}

\DeclareMathAlphabet{\mathtst}{OMS}{cmsy}{m}{n}

\DeclareFontEncoding{FML}{}{}%
\DeclareFontSubstitution{FML}{futm}{m}{it}%
\DeclareSymbolFont{fourier}{FML}{futm}{m}{it}%
\DeclareMathSymbol{\partialup}{\mathord}{fourier}{130} 

\definecolor{green}{rgb}{0.564706,0.933333,0.564706}
\definecolor{red}{rgb}{0.941176,0.501961,0.501961}

\usepackage{ifthen,version}
\newboolean{include-notes}
\setboolean{include-notes}{true}
\newcommand{\tmnote}[1]{\ifthenelse{\boolean{include-notes}}%
  {\textbf{(TM says: #1)}}{}}
\newcommand{\tmadd}[1]{\ifthenelse{\boolean{include-notes}}%
  {\textcolor{red}{#1}}{#1}}
\newcommand{\jsnote}[1]{\ifthenelse{\boolean{include-notes}}%
  {\textbf{(JS says: #1)}}{}}
\newcommand{\jsadd}[1]{\ifthenelse{\boolean{include-notes}}%
  {\textcolor{red}{#1}}{#1}}

\begin{document}
%
\title{Dynamic Task Execution using Active Parameter Identification with the Baxter Research Robot}
%
%
%

\author{Andrew D. Wilson,~\IEEEmembership{Member,~IEEE,}~Jarvis A. Schultz,~\IEEEmembership{Member,~IEEE,}\linebreak
       Alex R. Ansari,~\IEEEmembership{Member,~IEEE,} and~Todd D. Murphey,~\IEEEmembership{Member,~IEEE}
\thanks{This work was supported by the National Science Foundation under Grant CMMI 1334609. Any opinions, findings, and conclusions or recommendations expressed in this material are those of the author(s) and do not necessarily reflect the views of the National Science Foundation.}
\thanks{A. Wilson, J. Schultz, A. Ansari, and T. Murphey are with the Department of Mechanical Engineering,
        Northwestern University, 2145 Sheridan Road, Evanston, IL 60208, USA. 
        awilson@u.northwestern.edu, jschultz@northwestern.edu, alexanderansari2011@u.northwestern.edu, and t-murphey@northwestern.edu}%
\thanks{This is the accepted preprint version of the manuscript.  The final published version is available at http://dx.doi.org/10.1109/TASE.2016.2594147.}}%

\markboth{IEEE TRANSACTIONS ON AUTOMATION SCIENCE AND ENGINEERING}%
{Wilson \MakeLowercase{\textit{et al.}}: Dynamic Task Execution using Active Parameter Identification with the Baxter Research Robot}
%

\maketitle

\begin{abstract}
This paper presents experimental results from real-time parameter estimation of a system model and subsequent trajectory optimization for a dynamic task using the Baxter Research Robot from Rethink Robotics.  An active estimator maximizing Fisher information is used in real-time with a closed-loop, non-linear control technique known as Sequential Action Control. Baxter is tasked with estimating the length of a string connected to a load suspended from the gripper with a load cell providing the single source of feedback to the estimator. Following the active estimation, a trajectory is generated using the {\tt trep} software package that controls Baxter to dynamically swing a suspended load into a box. Several trials are presented with varying initial estimates showing that estimation is required to obtain adequate open-loop trajectories to complete the prescribed task. The result of one trial with and without the active estimation is also shown in the accompanying video.
\end{abstract}

\renewcommand{\abstractname}{Note to Practitioners}
\begin{abstract}
This paper experimentally demonstrates the capability of an on-line parameter learning algorithm on the Baxter Research Robot to improve task performance.  This type of algorithm could enable automated systems to actively inspect multi-body parts for parametric information including estimation of the robot's own inertias.  The method requires known equations of motion for any nonlinear system with uncertain, constant parameters.  We show using a series of 18 experimental trials that using the estimation method results in improved task performance for automated dynamical motions given uncertain parameters.
\end{abstract}


\section{Introduction}
%
\IEEEPARstart{O}{ne} fundamental goal of artificial learning for automation and production is providing the capability for a robot to automatically synthesize actions that improve estimates of the robot's internal dynamics and dynamic models of real-world objects.  Human workers on production lines constantly use dynamic interactions with objects to improve their quality and speed in a manufacturing environment. We aim to provide this form of learning on robots using real-time processing of feedback from active exploration of the environment. Since the general problem of model synthesis and learning remains a formidable one, we restrict ourselves in this paper to creating a method for real-time active synthesis of dynamic trajectories to estimate a single model parameter in a known dynamical model.  

\begin{figure}[t]
\centering
\includegraphics{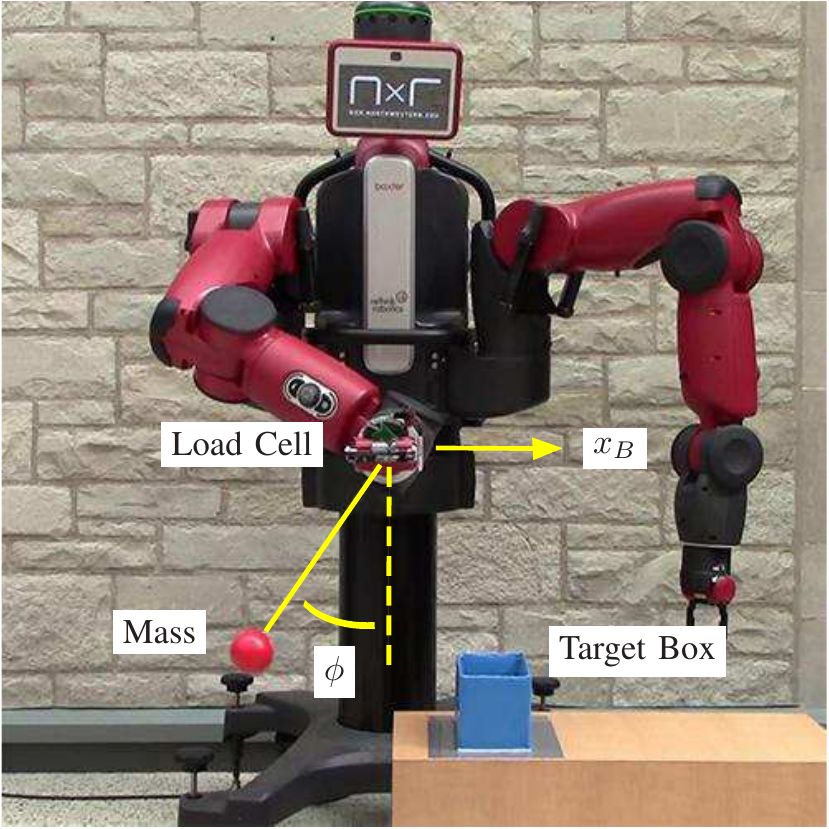}
\caption{Baxter performing real-time active parameter estimation.}
\label{fig:baxter}
\end{figure}

Active estimation of parameters within dynamical systems, also referred to as \textit{optimal experimental design}, commonly uses Fisher information as the primary metric \cite{Emery1998}. Fisher information provides a best-case estimate of the estimator's performance given a set of measurements from a robot through the Cramer-Rao bound \cite{Rao1947, Bar-Shalom2004}. A number of works on ``exciting'' trajectories by Armstrong and others  \cite{Armstrong1989, Swevers1997, Gautier1992} provide the theoretical basis for information-based estimation.  In work by Emery \cite{Emery1998},  least-squares and maximum-likelihood estimation techniques are combined with Fisher information to optimize the experimental trajectories.  In this case and several others, dynamics are solved as a discretized, constrained optimization problem \cite{Mehra1974,Vincent2010,Franceschini2008}. One downside is that this time discretization can lead to high dimensional optimization problems (dimensions of $10^7$ to $10^{12}$ are common in practice).

One question that may be raised by the reader is why a trajectory optimization algorithm is necessary for excitation. Non-algorithmic approaches such as frequency sweeps on the control inputs can be performed which will likely provide some level of Fisher information; however, the real cost to a sweep approach is a large use of energy in the control inputs. Using an optimization algorithm generates trajectories that provide an appropriate level of information with far less control energy than a non-optimal excitation. Especially for under-actuated systems, exciting certain harmonics of the system and exploiting the free dynamics is critical to minimizing the control energy which is achieved through trajectory optimization.

This paper expands on preliminary results by the authors using Sequential Action Control for Fisher information maximization and parameter estimation. The preliminary results, presented at the 2015 IEEE/RSJ International Conference on Intelligent Robots and Systems \cite{Wilson2015c} include the parameter estimation algorithm while this paper provides results from a practical task involving a dynamic system. Detailed derivations of the underlying control principles and information theory can be found in \cite{WilsonTRO, SACtro, AnsariRAL, Johnson2009}. The Baxter Research Robot, shown in Fig. 1,  has been used as a practical platform for a number of studies \cite{Caldwell2014, Mao2014, Bowen2014} while also presenting a number of challenges including high compliance and actuator saturation.  Despite these potential sources of unmodeled dynamic effects, results show that the estimator successfully converges to the actual parameter value across several trials and completes the dynamic task with the correct parameter estimates.  

This paper is organized as follows: Section \ref{sec:algorithm} provides an overview of the algorithmic foundation for the experiment. The specific implementation details for Baxter and dynamic task are provided in Section \ref{sec:implementation}. Section \ref{sec:results} provides results from several trials of the real-time estimation task, and Section \ref{sec:conclusion} concludes the paper with notes on future work.  The estimation and dynamical task execution presented in this paper are also shown in the accompanying video.

\section{Algorithmic Overview}
\label{sec:algorithm}

This section presents an overview of the optimal control algorithms implemented in Section \ref{sec:implementation}. There are two stages to the control problem: First, an unknown parameter must be estimated in the system. For the example presented in this paper, the string length of the suspended mass is uncertain. Acquiring a better estimate will allow the optimized task trajectory to complete successfully; however, estimation of the parameter requires active exploration by the robot. The active estimation algorithm involves an extension of the Sequential Action Control (SAC) algorithm \cite{SACtro, AnsariRAL}.

After estimation, the second stage is the synthesis of the task trajectory. The optimization of the task trajectory is performed using a projection-based nonlinear trajectory optimization routine provided by 
{\tt trep}, a simulation and optimal control package available at {\tt http://nxr.northwestern.edu/trep}. Since there is no state feedback in this example, the task trajectory is run open-loop which requires accurate model parameters obtained from the first stage. The following sections provide detail on the two control stages.

\subsection{Active Parameter Estimation}
\label{sec:active-estimation}
\begin{figure}[t]
  \centering
  \includegraphics[trim = 2.2in 0.6in 2.2in 0.6in, clip, width=3.2in]{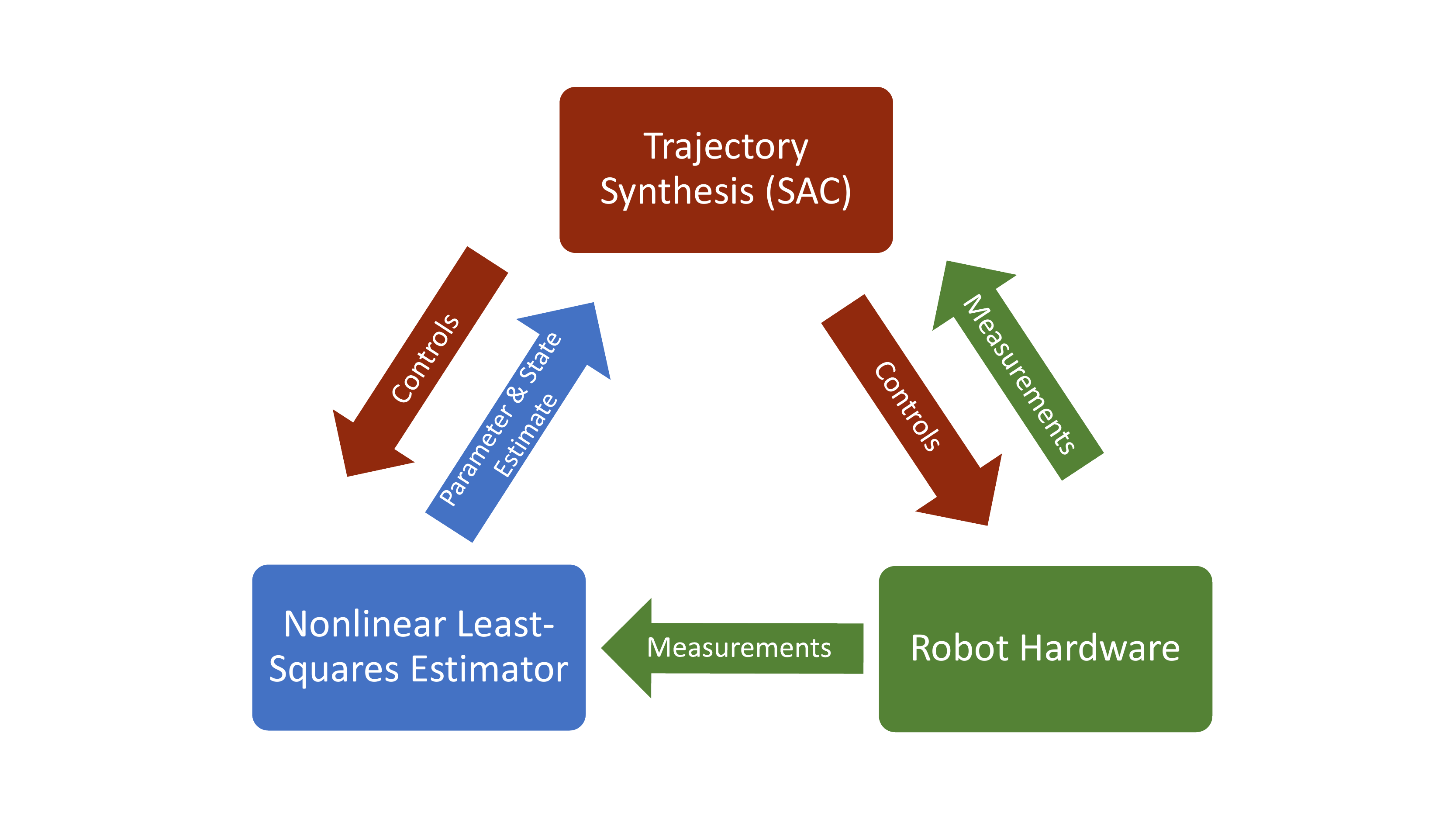}
  \caption{Overview of the SAC active estimator real-time control structure.}
\label{fig:control-flow}
\end{figure}

As shown in Fig. \ref {fig:control-flow}, there are two primary modules interacting with the robot hardware: the SAC controller which synthesizes a trajectory that locally maximizes Fisher information and the nonlinear least-squares estimator which provides on-line updates to the current best estimate of the uncertain parameter set.  At the highest level, the least-squares estimator requires the control inputs provided by the SAC controller to compute a predicted output which is compared to the actual measurements provided by the robot.  The SAC controller is updated with new parameter estimates and state estimates by the least-squares estimator and optionally can receive state feedback directly from the robot hardware if available. These modules can be run asynchronously and at different rates with the use of a nonlinear state observer model.

\subsubsection{SAC Trajectory Synthesis}
In this paper, we assume that one parameter is uncertain with additive noise on observer measurements but negligible process noise.  The same cost function with several unknown parameters is presented in \cite{WilsonTRO}. For SAC, the state is usually derived assuming control-affine dynamics \cite{SACtro, AnsariRAL}. Thus, the model of the system is defined as 
\begin{align}
\dot{x} =& f(x,u,\theta) = g(x, \theta) + h(x, \theta) u
\label{eq:dynamics}\\
\tilde{y}=&y(x,u,\theta)+\mathrm{w}_y\nonumber
\end{align}
where $x\in\mathbb{R}^n$ defines the system states, $\tilde{y}\in\mathbb{R}^h$ defines the measured outputs, $u\in\mathbb{R}^m$ defines the inputs to the system, $\theta\in\mathbb{R}$ defines the parameter to be estimated, and $\mathrm{w}_y$ is additive output noise where $p(\mathrm{w}_y)=N(0,\Sigma)$. 

In order to maximize information, the SAC cost function is modified to include a cost on the Fisher information of the uncertain parameter. For this implementation, we assume that the measurement noise of the system is
normally distributed with zero process noise. Therefore the Fisher information is given by
\begin{equation}
I(\theta)=\sum_{k=k_0}^{k_f}\Gamma_\theta(t_k)^T\cdot \Sigma^{-1}\cdot \Gamma_\theta(t_k)
\label{eq:information}
\end{equation}
where $\Gamma_\theta$ is the derivative of the output $y$ w.r.t. the parameter $\theta$ given by
\begin{align*}
\Gamma_\theta(t_i)=&D_xy(x(t_i),u(t_i),\theta) \cdot D_\theta x(x(t_i),u(t_i),\theta)\nonumber\\
&+D_\theta y(x(t_i),u(t_i),\theta).
\end{align*}

As detailed in \cite{WilsonTRO}, a cost function on Fisher information from (\ref{eq:information}) requires the simulation of the gradient of $x$ w.r.t. $\theta$, i.e. $\psi(t) = D_\theta x$. These additional states are referred to in the paper as extended state dynamics and notated as $\bar{x}=(x, \psi)$. For this implementation, we use only a running cost for the Sequential Action Controller which is written as 
\begin{equation}
\label{J}
J_{\tau} = \int_{t_0}^{t_f} l(\bar{x}(t)) \,dt
\end{equation}
where
\begin{equation*}
l(\bar{x}(t)) = \left[\Gamma_\theta(t)^T\cdot \Sigma^{-1}\cdot \Gamma_\theta(t)\right]^{-1}+ x(t)^T \cdot Q_{\tau} \cdot x(t)
\end{equation*}
i.e., the minimization of the inverse of the information and an optional trajectory tracking cost to bias the system toward a particular part of the state space.

\begin{figure}[t]
  \centering
  \includegraphics[width=3.3in]{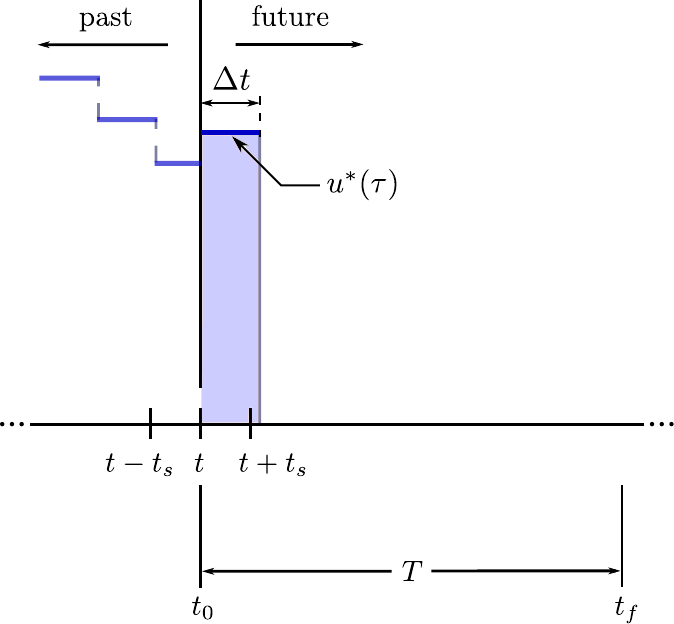}
  \caption{SAC actions for a 1-D control are sequenced in receding-horizon fashion.}
\label{fig:sac_action}
\end{figure}

The SAC control synthesis process follows a receding-horizon style format to sequence together separately short optimal control \emph{actions} into a piecewise continuous constrained feedback response to state, similar to a nonlinear model predictive controller but with a single control action \cite{camacho2013,grune2011}. In SAC, \emph{actions} are defined by a pair composed of a control vector value and its associated (typ. short) application duration.  The blue shaded region in Fig.~\ref{fig:sac_action} shows a SAC action for a 1-D control where $u^*(\tau)$ is the control and $\Delta t$ is the duration.

The SAC algorithm predicts system motion from current state feedback, $x(t_0) = x_0$. The process involves simulation of a state and adjoint system $(x,\rho)$ for a fixed time horizon, $T$, until (receding) final time $t_f = T + t_0$. For the purposes of this paper, the nominal control value used for simulations $(x,\rho)$ is $u = 0$ so that SAC computes optimal actions relative to the free (unforced) system motion.

The adjoint variable $\rho : \mathbb{R}^{} \mapsto \mathbb{R}^{2n}$ provides information about the sensitivity of the cost function to the extended state, $\bar{x}$.  The algorithm maps this sensitivity to a control sensitivity provided by an inner product between the adjoint and dynamics \eqref{eq:dynamics}. The process of control synthesis uses this sensitivity, $\frac{dJ_{\tau}}{du_{\tau}}$, to search for least norm actions that optimize the expected change in cost \eqref{J}. Thus SAC actions optimize the rate of trajectory improvement. These optimal actions depend directly on the adjoint, which is determined from open-loop simulation of the following equation,
\begin{equation*}
\dot \rho = -D_{\bar{x}} l(\bar{x})^T - D_{\bar{x}} f(\bar{x},u)^T \rho
\end{equation*}
with a terminal condition $\rho(t_f) = 0$. The adjoint equation is evaluated along the nominal system trajectory with $u=0$.  The ability to calculate an optimal action at the current time from a single adjoint differential equation enables the use of the algorithm for real-time computation.  For a complete derivation of SAC control synthesis with examples see \cite{SACtro}.

\subsubsection{Nonlinear Least-Squares Estimator}
While the robot is executing a motion, a nonlinear least-squares estimator is used on-line to update the estimated value of the parameter as well as the robot state. The least-squares estimator can be written as
\begin{eqnarray}
\hat{\theta} = \arg\min_{\theta} \beta(\theta) \label{eq-parameterestimate}
\end{eqnarray}
where
\begin{equation}
\label{eq:leastsquares}
\beta(\theta)=\frac{1}{2}\sum\limits_{i}^{h}(\tilde{y}(t_i)-y(t_i))^T\cdot\Sigma^{-1}\cdot(\tilde{y}(t_i)-y(t_i)).
\end{equation}
$\tilde{y}(t_i)$ is the observed state at the $i^{th}$ index of $h$ measurements, $\Sigma\in\mathbb{R}^{h\times h}$ is the covariance matrix associated with the sensor measurement error, and $\hat{\theta}$ is the least-squares estimate of the parameter. The estimator recomputes a new estimate at a set frequency, incorporating any new measurements received since the last iteration.

Given this estimator, we will use gradient descent with a backtracking line-search to find optimal parameter values by minimizing the least-squares error in (\ref{eq:leastsquares}). Since the estimator requires a predicted output $y(t)$, which is based on current estimates of $\theta$ and $x(t)$, a nonlinear state observer is also required for systems without full-state feedback.  While any numerical differential equation solver may be sufficient, for this implementation, we use the {\tt trep} software package, which is also used for the task trajectory optimization detailed in the following section.

\subsection{Task Trajectory Optimization}
Once the uncertain parameter has been accurately identified using the SAC active estimator, the task trajectory must be synthesized.  At the end of the estimation process, Baxter's end effector is returned to the zero position and the motion of the mass is allowed to naturally dampen while the task trajectory computation completes.  This process can be seen in the accompanying video.

The selection of an appropriate optimal control method is dependent both on the computational requirements and level of dynamic motion in the task. Sequential Action Control was selected for the on-line estimator since real-time performance was desired. To achieve real-time performance, the SAC algorithm computes the current control over a long time horizon; however, future controls are not taken into account when computing an optimal action.  For the estimator, this is acceptable since the model parameters are incorrect, and future controls are largely irrelevant relative to the measurement feedback. Once the parameters are known, a complete trajectory optimization method can then be used.

For this paper, optimization of the task trajectory is completed using the {\tt trep} software package which implements a discrete mechanics version of projection-based nonlinear optimal control \cite{schultz_cism_2015, Hauser2002}. This method provides a trajectory that is locally optimal w.r.t. the initial trajectory provided. For this example, a box is placed 0.45m to the left of the suspended mass.  To slightly simplify the optimization method, the dynamic task is to swing the mass to a fixed point over the box with zero velocity at the terminal point. At the end of the optimized trajectory, the end effector is quickly moved directly over the box, allowing the mass to free-fall into the box. This action can be seen in the results from the accompanying video. The initial trajectory for the optimization routine is set as the stationary trajectory with zero controls.

The objective function for the task trajectory uses a terminal cost on the nominal states of the suspended mass and a running cost on the control given by
\begin{align}
J_{\mathrm{task}}=&(x(t_f)-x_d(t_f))^T\cdot P_\tau \cdot (x(t_f)-x_d(t_f))\nonumber\\
&+\int\limits_{t_0}^{t_f}u(t)^T\cdot R_\tau \cdot u(t)\hspace{0.03in} dt,
\label{eq:trajcostcont}
\end{align} 
where $P_\tau$ and $R_\tau$ are the state and control weights. 

Each step of the projection-based optimization algorithm returns a new dynamically feasible trajectory with an improved cost.  The software uses discrete-time algebraic Riccati equations resulting from a LQ problem formulation to produce a perturbation to the current trajectory which is then projected to a feasible curve. This process is repeated iteratively until the magnitude of the cost derivative is below a specified tolerance.  For the results presented in this paper, the tolerance is set to $10^{-6}$.

\section{Baxter Implementation}
\label{sec:implementation}

This section describes both the problem formulation and software specific
implementation of the control structure described in the previous section on the
Baxter Research Robot created by Rethink Robotics
\cite{baxter2015}. Results from experimental trials are presented in Section IV.

\subsection{Problem Description and Model Formulation}
\label{sec:probl-descr-model}

\begin{figure*}[t]
	\centering
	\subfloat[Incorrect parameter plan failing to swing the mass into the box.]{
		\includegraphics[height=2.25in]{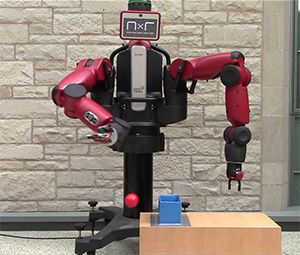}}\hspace{0.25in}
	\subfloat[Correct parameter plan successfully swinging the mass into the box.]{
		\includegraphics[height=2.25in]{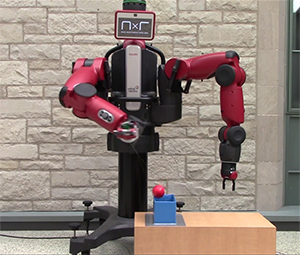}}
	\caption{Baxter near the completion of the dynamic task for both incorrect and correct parameter plans.}
	\vspace{-0.1in}
	\label{fig:baxter-task}
\end{figure*}

To test the learning capabilities on a practical robotic system, we
created a dynamical task which involves swinging a suspended mass into a nearby box. 
In order to require the use of dynamics to solve the problem, we restrict Baxter's end effector in software to only move along the horizontal axis $x_B$; therefore, the mass must be swung to land inside the box shown in Fig. \ref{fig:baxter-task}. However, we assume that the string length of the suspended mass is uncertain and must be estimated prior to optimizing the task trajectory.

As there is no feedback on the angle of the suspended mass, a load cell mounted at Baxter's end effector is used as the sole measurement input to the estimator. While load measurements on the end effector could be made using the joint torque sensors on the Baxter platform, the measurement noise is significantly greater and non-Gaussian due to the series-elastic actuation which degrades the effectiveness of the estimator.

This configuration necessitates active motion to estimate the string length as the Fisher information is zero when the robot is stationary. To simplify the control of Baxter, we chose to approximate the end effector as kinematic and control motion only in the Cartesian x-axis, $x_B$.
The equations of motion for the active estimation stage are given by the following,
\begin{equation*}
f_{\mathrm{SAC}}(x,u,\theta)=\left[\begin{array}{c}
\dot{x}_B\\
u\\
\dot{\phi}\\
\frac{u}{\ell}\cos \phi - \frac{g}{\ell} \sin \phi
\end{array}\right]
\end{equation*}
where $u$ is the x-axis acceleration of the gripper, $m$ is the known mass suspended from the robot, and $\ell$ is the length of the string, which will be estimated. 
Additionally, the equation for the force output $F_s$ is
\begin{equation*}
y(x,u,\theta)=F_s=m g \cos\phi - m \ell \dot{\phi}^2-u\sin\phi.
\end{equation*}
It is assumed that the trajectories will maintain tension in the string; therefore, the distance between the robot and mass is fixed.  Given this system model, the extended state $\bar{x}\in\mathbb{R}^8$.

After the first estimation stage, a slightly different choice of states is made for the task trajectory in the second control stage. Since a terminal cost is given as a function of the Cartesian position of the suspended mass in the vertical $x$-$z$ plane, those will be the system states along with the position of Baxter's end effector $x_B$. A distance constraint is added in the {\tt trep} software which defines a constant distance between the end effector and mass which represents the string length. 

\subsection{Experimental Implementation}
\label{sec:exper-impl}
As shown in Fig. \ref{fig:control-flow}, there are essentially four modules
interacting with the robot hardware: the SAC trajectory synthesis module, a measurement module,
the nonlinear least-squares estimator module, and the task trajectory module. In this section we describe how
each of these modules is implemented for the Baxter experiments.

The backbone of communication between modules and the robot is provided by the Robot Operating System (ROS) \cite{quigley2009ros}.
ROS provides the ability to asynchronously run the control and estimation modules, implemented as ROS nodes, and
Baxter natively uses ROS as the primary API for motion commands. Three primary nodes are employed: the SAC control node, 
a measurement receiver node, and the estimator node.

\subsubsection{Baxter SAC Node}
\label{sec:baxter-sac}

As discussed in the previous section, the end effector of the Baxter
robot is approximated as a kinematic input to the dynamic suspended mass
system. The dynamic model for the suspended mass system assumes only planar
motion, and the kinematic input moves perpendicular to gravity in this plane.
It follows that the SAC module produces target locations along a one-dimensional
line that Baxter's end effector should follow. A joint velocity controller for
Baxter's right arm is used to stabilize the right end effector to these target
locations. To avoid issues with kinematic singularities in the manipulator while
controlling to these target locations, this one-dimensional line is expanded to
a candidate set of closely-spaced end effector targets in $SE(3)$. An off-line
computation is done to solve the inverse kinematics problem for each of the
targets in the set producing a set of target joint angles for each target. These joint angle targets
are then stored to disk as a lookup table between target horizontal positions,
like those produced by SAC, and target joint angles. During an experimental run,
the desired joint position, velocity, and acceleration data is sent using a Joint Trajectory Publisher to Baxter's internal controller which runs a high frequency real-time loop to control each of the joints.

\subsubsection{Baxter Measurement Receiver Node}
The payload is attached to Baxter's end effector through a one-dimensional shear load cell from Phidgets with a 100g weight capacity. A microcontroller with a HX711 chip for A/D conversion samples the load cell at 100~Hz and transmits the
measured forces via a serial link back to a client computer which is communicating with Baxter 
over an Ethernet connection. These force measurements represent the $\tilde{y}(t_{i})$ in (\ref{eq:leastsquares}).  The load 
cell has been calibrated prior to the experiment to convert the load cell output
to a force (N).  The resulting force is timestamped as it is received and published at 100 Hz for use by the estimator node. 

\subsubsection{Baxter Estimator Node}
\label{sec:baxt-estim-module}
The estimator node subscribes to the actual end effector trajectory which is provided by the Baxter API. These measurements are used to generate the $y(t_{i})$ terms in (\ref{eq:leastsquares}) through the use of a nonlinear state observer.  As mentioned in Section \ref{sec:active-estimation}, the {\tt trep} software package is used as the state observer.  The trajectory is used as the input and {\tt trep} provides the predicted state evolution of the suspended load. From these states, the predicted force, $y(t)$ can be calculated.

At a frequency of 2~Hz, the estimation module solves the optimization problem
described by (\ref{eq-parameterestimate}), updating the estimate of the string length. This frequency was chosen as a conservative rate at which the estimator has enough time to provide an update; however, the rate can be modified depending on the required computational time for the particular system.  This parameter estimate and new state estimate are provided as a service to the SAC node which queries the service at the start of each computation. Updates to the string length estimate and expected state are reflected in this service call.

\subsubsection{Baxter Task Node}
The previous three nodes are all used for the estimator in the first control stage. Following the estimation, a task node is created that uses {\tt trep} to compute an optimal task trajectory using the estimated value of the string length.  {\tt trep} uses a discrete timestep of 0.01s for the numerical simulation resulting in a control frequency of 100Hz for the task execution.  

Position, velocities, and accelerations are sent open-loop to the Baxter robot using a Joint Trajectory Publisher in ROS. The same lookup table as in the SAC node is used to generate joint positions and a corresponding Jacobian table is used for velocities.  Accelerations are provided using a finite difference of the joint velocities. A finite-time horizon of 5.0 s is used for the task trajectory. At the end of the trajectory, the end effector is commanded to a position over the box that allows the suspended load to fall directly into the box.

\section{Experimental Results}
\label{sec:results}

This section presents the results of experimental trials of the active estimation algorithm using the Baxter robot. To compare the ultimate goal of task performance given uncertain initial estimates of the string length of the suspended mass, a total of 18 trials were run. The first set of 9 trials were run with varying initial length estimates which are then used directly in the task trajectory optimization without the active estimation stage. The second set of 9 trials were run using the same distribution of initial estimates, but the active estimation stage was also used to improve the values of the string length for the task optimization.

\subsection{Direct Task Optimization Results}
To evaluate the performance of the Baxter robot with an open-loop task and uncertain parametric information, the first 9 trials were run using the initial length estimates given in the first column of Table \ref{tab:taskresults}. Using {\tt trep}, a trajectory was synthesized using a parameter value from the table and the task was attempted. If the suspended mass landed inside the box at the end of the trial, the trial is considered a success. Any trial with the mass outside of the box at the end of the trial resulted in failure.

\begin{figure}[t]
	\centering
	\includegraphics{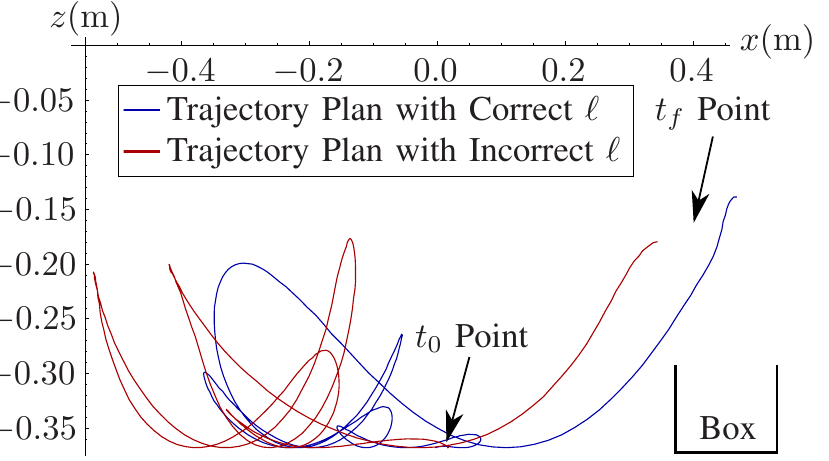}
	\caption{Simulated trajectory plans for the suspended mass with the correct $\ell=0.368$m and incorrect $\ell=0.328$m.}
	\vspace{-0.1in}
	\label{fig:xzplot}
\end{figure}

The results of these 9 trials are shown in the center column of Table \ref{tab:taskresults}. The task is completed with only two estimates of the length - the actual value of 0.368m and a slightly longer length of 0.388m. It is important to note that this experiment only provides a single task result at each value. Therefore, it is likely that as the parameter deviates from the actual value, the task may or may not be completed with some distribution at each value; however, this experiment provides a qualitative pattern of results when varying the initial parameter estimate.  Figure \ref{fig:xzplot} shows the difference between one path of the suspended mass simulated at correct value of 0.368m and the simulated path planned at an incorrect value of 0.328m. As shown in the figure, the incorrect parameter value results in motion which does not swing the mass far enough over to land above the box.   
This difference is also highlighted in frame captures from the experimental trials in Fig. \ref{fig:baxter-task}. The experimental frame captures mirror the end of the simulated trajectories from Fig. \ref{fig:xzplot} with the incorrectly planned trajectory failing to swing the suspended mass over the top of the box.

In order to improve the performance of this open-loop task given the uncertain length, the length estimate must be improved using the active estimation algorithm prior to attempting the task.

\subsection{Task Results with SAC Active Estimation}
The second set of 9 trials were run with the first stage SAC estimator followed by the same task trajectory synthesis stage used in the previous set of trials. However, in this set of trials, the parameter value is identified using the estimator and the result is used in the task trajectory stage. 

\begin{figure}[t]
	\centering
	\includegraphics{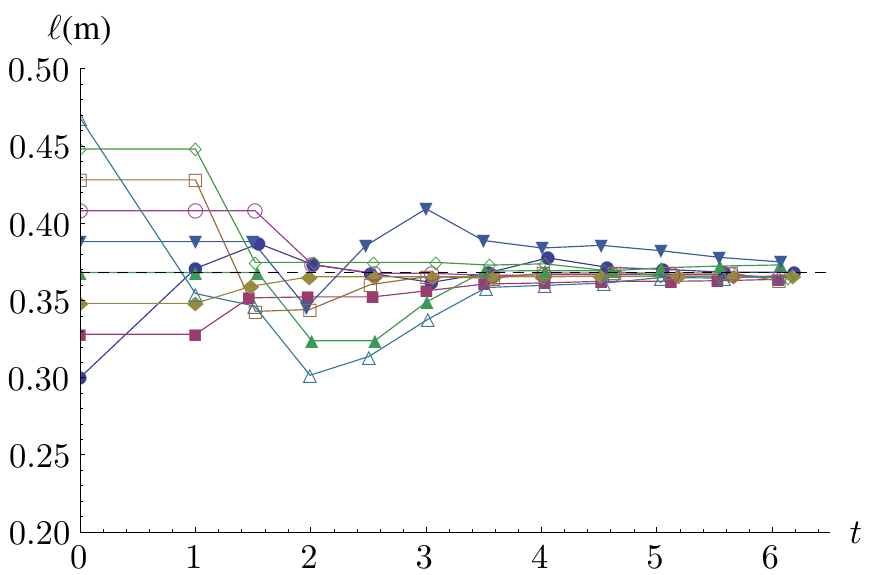}
	\caption{Experimental trials on Baxter using different initial estimates of the string length.  The dashed line indicates the actual measured length.}
	\label{fig:baxter-initexp}
	\vspace{-0.05in}
\end{figure}

The resulting parameter values for the 9 trials throughout the active estimation stage are shown in Fig. \ref{fig:baxter-initexp}. Estimation begins after the first second and the parameter values are updated as more information is acquired through the load cell measurements.  Since the estimator uses a sufficient decrease condition on the least-squares problem, the value is only updated when the sufficient decrease is satisfied.

\begin{table}[t]
	\caption{Experimental Task Results}
	\label{tab:taskresults}
	\centering
	\begin{tabular}{c c c}
		\hline\hline
		Initial Length&Without SAC&With SAC\\
		Estimate (m):&Estimation:&Estimation:\\
		0.308&\cellcolor{red}Fail&\cellcolor{green}Success\\
		0.328&\cellcolor{red}Fail&\cellcolor{green}Success\\
		0.348&\cellcolor{red}Fail&\cellcolor{green}Success\\
		\hspace{0.05in}0.368*&\cellcolor{green}Success&\cellcolor{green}Success\\
		0.388&\cellcolor{green}Success&\cellcolor{green}Success\\
		0.408&\cellcolor{red}Fail&\cellcolor{green}Success\\
		0.428&\cellcolor{red}Fail&\cellcolor{green}Success\\
		0.448&\cellcolor{red}Fail&\cellcolor{green}Success\\
		0.468&\cellcolor{red}Fail&\cellcolor{green}Success\\
		\hline\hline
		*actual string length
	\end{tabular}
	\vspace{-0.1in}
\end{table}

The estimates clearly converge toward the actual string length, which is noted by the dashed horizontal line.  Qualitatively, the rate of convergence across all the trials is similar, with estimates beginning to converge around 2 to 4 seconds.  This suggests that for this system, the trajectories generated by the SAC algorithm provide relatively similar levels of information despite the initial estimate.  Since the estimator cost may not be convex, initial estimates too far from the actual value may not converge to the true value due to the presence of local minima. The mean estimate of $\ell$ from the 9 trials after 6 seconds was 0.367m with the actual string length set to 0.368m.  The standard deviation of the final estimates is 0.0042m.

\begin{figure}[t]
	\centering
	\includegraphics{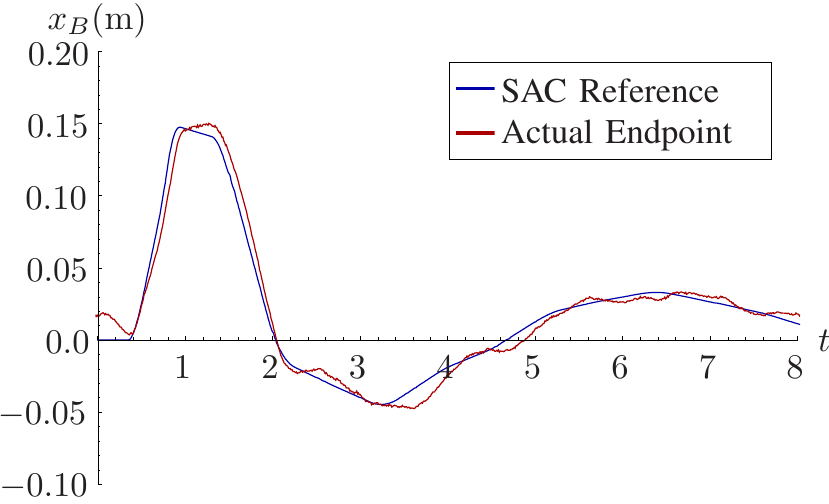}
	\caption{Endpoint reference compared to actual endpoint position of the Baxter trial with $\ell_0=0.45$ m.}
	\label{fig:endpoint}
	\vspace{-0.1in}
\end{figure}

The generated and executed trajectories can be seen for one of the trials in Fig. \ref{fig:endpoint}.  As discussed in Section \ref{sec:implementation}, the motion of the gripper is controlled to move along the Cartesian x-axis using PID control to follow the generated reference. The use of the Joint Trajectory Publisher along with the Baxter API results in reasonable tracking performance given the dynamic nature of the desired motion.

Immediately following the estimation stage for each trial, the task trajectory is synthesized and the task is attempted.  The results of the task stage can be seen in the third column of Table \ref{tab:taskresults}. Since the estimator has accurately identified the previously uncertain string length, the task is successfully completed from each initial length estimate for all 9 trials. For a more complete view of the experiment, the accompanying video shows one complete trial including the task trajectory optimization.


\section{Conclusion}
\label{sec:conclusion}
This paper presented an experimental implementation of control for real-time active estimation to improve open-loop task performance. Results from several trials on the Baxter robot with different initial estimates of the string length quickly converge to the actual length using a trajectory synthesized on-line using a Sequential Action Controller. This improvement in the parameter estimate allows a dynamic task to be completed despite a distribution of initial estimates of the parameter.

This work represents only one step toward improving robot learning on physical systems by exploiting dynamic models. One possible improvement may include the use of Lie groups as the fundamental tool to build the dynamic models.  While the algorithms used to evaluate the Lie group models may be more complicated, they can facilitate better-posed solutions to the optimization problems, reducing singularities and angle wrapping issues common in robotic systems, especially non-holonomic systems.

Additionally, the extension to multiple parameters only requires that an optimality metric is set as shown in \cite{WilsonTRO}; however, it would be useful for a system to realize which parameters need to be estimated in the first place.  This could be achieved through forms of sensor fusion and covariance estimation or through exploration-based search algorithms.  Eventually, the addition of a model creation and learning algorithm would enable a robot to develop not only estimates of the parameters, but also internal structure without prior knowledge of the internal dynamic model.  The continued development of dynamics-based methods for robot learning will allow robots to learn and better interact with real-world objects and tasks in physical environments.

\bibliographystyle{ieeetr}
\bibliography{TASE_short}

\end{document}